\documentclass[letterpaper, 10 pt, conference]{ieeeconf}  

\IEEEoverridecommandlockouts                              

\usepackage{amsmath, amssymb, bbm, setspace, mathrsfs, color, listings, graphics, graphicx, multicol, dcolumn, dsfont, subcaption, float, mathabx, hhline}
\usepackage[font=footnotesize,labelfont=bf,skip=5pt]{caption}
\usepackage{multirow}
\usepackage{hyperref}
\usepackage{breakurl}

\newcommand{\paren}[1]{\left( #1 \right)}

\newcommand\norm[1]{\left\lVert#1\right\rVert}

\title{\LARGE \bf
CAZSL: Zero-Shot Regression for Pushing Models \\by Generalizing Through Context
}

\author{Wenyu Zhang$^{1}$, Skyler Seto$^{1}$ and Devesh K. Jha$^{2}$%
\thanks{$^{1}$Wenyu Zhang and Skyler Seto are with Cornell University Department of Statistics and Data Science {\tt\small \{wz258,ss3349\}@cornell.edu}}%
\thanks{$^{2}$Devesh K. Jha is with Mitsubishi Electric Research Laboratories (MERL), Cambridge, MA, USA 02139 {\tt\small jha@merl.com}}}%

\begin{document}

\maketitle
\thispagestyle{empty}
\pagestyle{empty}

\begin{abstract}

Learning accurate models of the physical world is required for a lot of robotic manipulation tasks. However, during manipulation, robots are expected to interact with unknown workpieces so that building predictive models which can generalize over a number of these objects is highly desirable. In this paper, we study the problem of designing deep learning agents which can generalize their models of the physical world by building context-aware learning models. The purpose of these agents is to quickly adapt and/or generalize their notion of physics of interaction in the real world based on certain features about the interacting objects that provide different \textit{contexts} to the predictive models. With this motivation, we present context-aware zero shot learning (CAZSL, pronounced as \textit{casual}) models, an approach utilizing a Siamese network architecture, embedding space masking and regularization based on context variables which allows us to learn a model that can generalize to different parameters or features of the interacting objects. We test our proposed learning algorithm on the recently released \textit{Omnipush} datatset that allows testing of meta-learning capabilities using low-dimensional data. Codes for CAZSL are available at \url{https://www.merl.com/research/license/CAZSL}.

\end{abstract}


\section{Introduction}

\begin{figure}[t]
	\centering
	\includegraphics[width=0.95\linewidth]{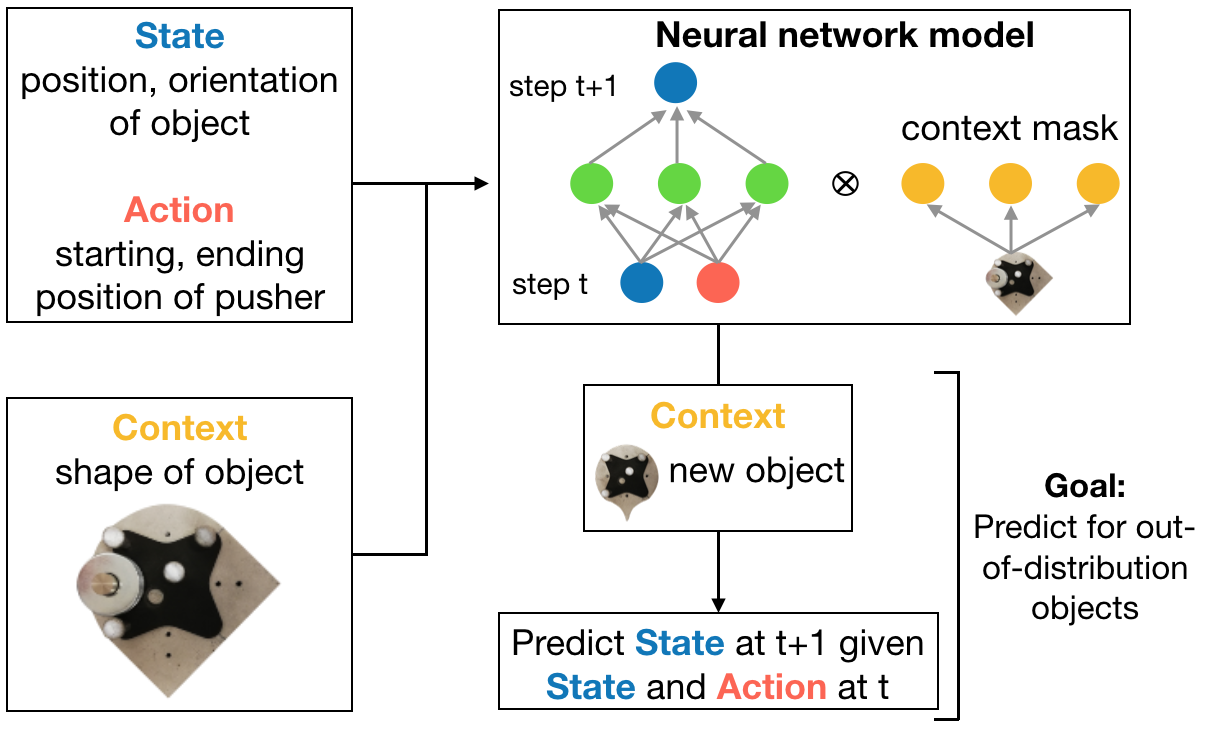}
	\caption{The proposed idea of learning context-aware zero-shot regression models in the paper. The \textit{context} variables are the additional features which effect the interaction dynamics being considered. The goal is for the learning agent can generalize to different context variables using the proposed approach.}
	\label{fig:introduction}
\end{figure}

\begin{figure}[t]
	\centering
	\begin{subfigure}{0.3\linewidth}
	    \includegraphics[height=\linewidth]{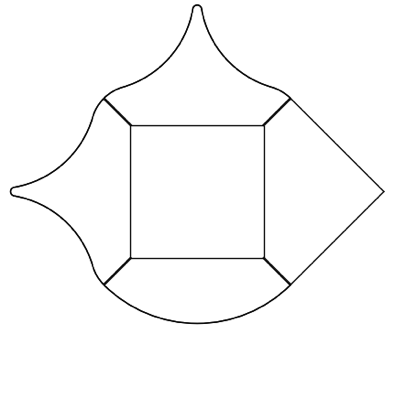}
	    \caption{\label{fig:obj0}}
    \end{subfigure}
	\begin{subfigure}{0.3\linewidth}
	    \includegraphics[height=\linewidth]{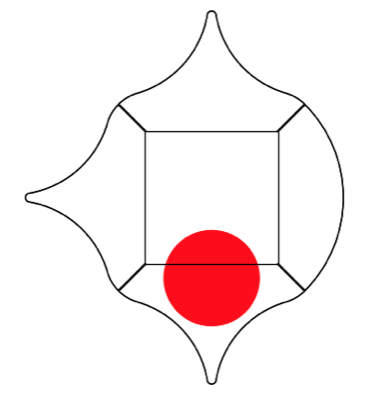}
	    \caption{\label{fig:obj1}}
    \end{subfigure}
	\begin{subfigure}{0.3\linewidth}
	    \includegraphics[height=\linewidth]{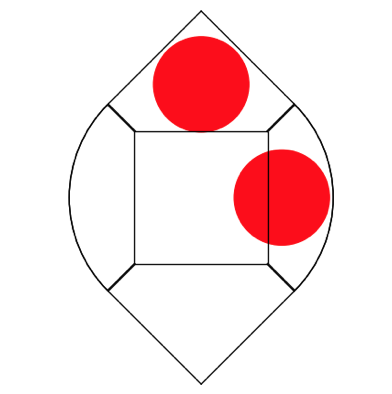}
	    \caption{\label{fig:obj2}}
    \end{subfigure}

  \vspace{-6pt}
	\caption{Three outlined top-down views of Omnipush objects with different shapes and weights. Red circles inside the object indicate the positions of weights. As explained in~\cite{bauza2019}, the pushing dynamics depends on the mass and shape of the object. It is desirable that a learning agent can quickly adapt its notion of pushing interaction based on these attributes of the objects. These attributes of a novel object can be obtained from an auxiliary system (e.g., a vision system). Images are reproduced from \url{http://web.mit.edu/mcube/omnipush-dataset/} with permission from authors~\cite{bauza2019}.}
	\label{fig:data_sample}
\end{figure}

Designing learning agents that can reliably perform robotic manipulation tasks is challenging~\cite{mason2018toward}. One of the reasons among many others is that robotic manipulation deals with a lot of challenging phenomena such as unilateral contacts, frictional contacts, impact, and deformation. These phenomena are challenging to understand or model even when considered individually, and manipulation requires considering several of these simultaneously. Consequently, it is difficult to either derive or learn precise models of interaction that can model different robotic manipulation tasks.
Furthermore, robots are expected to interact with unknown workpieces so that building predictive models which can generalize over objects is highly desirable and of practical value~\cite{mason2018toward}. For example, Figure~\ref{fig:data_sample} shows objects from the Omnipush dataset~\cite{bauza2019} where the pushing dynamics depend on the shape and mass distribution of the objects being pushed. While humans generalize effortlessly to variation in different physical properties of objects during interaction, it is difficult for robots to understand this generalization during interaction~\cite{battaglia2013simulation, smith2018different, osiurak2018looking, lake2015human}. 

Learning accurate models of the physical world is pre-requisite for many model-based robotic manipulation tasks. The motivation of our work is to train general purpose AI agents that can adapt their model of physical systems (e.g., interaction) using some extra features which can be easily obtained using auxiliary systems. For example, the interaction dynamics between two objects can depend on their mass, shape, size, etc. These features for a new object can be estimated using state-of-the-art vision systems or encoded into state-representation features~\cite{review}. The learning agents can then adapt their notion of the interaction physics based on these additional inputs. This is very similar to how humans adapt their model of objects based on features that they can sense. Throughout the paper, we call these additional features as \textit{context}. We propose zero-shot regression models outlined pictorially in Figure~\ref{fig:introduction} which are neural networks trained using these additional context variables. The concepts of meta-learning and zero-shot learning are very popular in machine learning literature~\cite{finn2017model}, and are increasingly applied in robotics~\cite{review,metaworld}. For the pushing application, while meta-learning requires support samples from unseen objects for adaptation~\cite{bauza2019}, we do not assume these new samples to be available in our work.

Supervised deep learning models are increasingly popular to model complex relationships in physical systems~\cite{sanchezgonzalez2018, ajay2018augmenting}.
The advantage of deep learning models lies in their superior ability to learn complex, non-linear spatial and temporal behaviors through the choice of large network architectures, which can then be optimized using large amounts of data. 

However in real applications, we are often unable to collect comprehensive datasets that cover all possible contexts, states and actions. 
For instance, we may be able to conduct physical experiments with a range of initial conditions for data collection, but not able to observe for all possible initial conditions.
Inductive biases typically allow deep learning models to generalize well to further samples collected under the same contexts. This renders such models suitable for applications with a finite and fixed set of contexts.
However, they may fare poorly with out-of-distribution samples from unseen contexts due to the lack of ability to generalize across contexts~\cite{kim2019attentive}, and hence need additional procedures such as context identifiers to correct for this~\cite{sanchezgonzalez2018}.

Inspired by these problems, we present a context-aware zero-shot learning method, CAZSL, for learning predictive models that can generalize across object-dependent context variables. We present a novel combination of context-based mask and regularizer that augments model parameters based on contexts and constrains the zero-shot model to predict similar behavior based on similarity in contexts. This allows us to make accurate prediction on novel objects by adapting the model based on the newly available context. The results of the proposed CAZSL method is reported using the Omnipush dataset which provides a diverse dataset with different objects for pushing dynamics. The proposed idea is presented pictorially in Figure~\ref{fig:introduction}, which presents the idea of CAZSL for the Omnipush dataset to generalize over the shape of objects for pushing. We demonstrate empirically that CAZSL improves performance or performs comparably to meta-learning and baselines methods in numerous scenarios.

\textbf{Contributions:} The proposed paper has the following contributions.
\begin{enumerate}
    \item We present a context-aware zero-shot learning (CAZSL) modeling approach with the motivation of building agents that can quickly adapt their notion of physics based on object-dependent context (analogous to parametric representation). We propose a novel combination of context mask and regularizer to constrain the model using similarities between contexts.
    \item We compare the proposed algorithm against several others methods on the recently released Omnipush dataset~\cite{bauza2019} providing new benchmark results for generalization. 
\end{enumerate}

Note that this paper only shows results for modeling using the proposed zero-shot learning approach. Use of the proposed models for model-based control is deferred to a future publication.

\section{Related Work}\label{sec:related_work}
The work presented in this paper is mainly motivated by the goal of creating generalizable models for learning complex interaction dynamics. These kind of physical interactions are common in a lot of physical systems. Interaction between objects especially plays a big role in robotic manipulation where a robot interacts with its environment using selective contacts~\cite{mason2018toward}. Learning accurate predictive models of physical systems and interactions is a very active area of research in robotics and machine learning~\cite{kober2013reinforcement}.

Model learning has been studied extensively in both machine learning as well as robotics community. The goal of these techniques in robotic manipulation is to learn high-precision models of interaction of the robot with the physical world which can be then used for synthesizing controllers~\cite{8794229, libera2020modelbased, finn2017deep, agrawal2016learning, Fazelieaav3123}.  Among the possible ways
to  manipulate  an  object,  pushing  stands  out  as  one  of  the most fundamental. As such, it has gained a lot of attention and thus, has been extensively studied~\cite{mason1986mechanics, lynch1992manipulation, lynch1996stable, 9297}. However, creating reliable models for pushing requires good models of friction, contacts, etc. which still remains poorly modeled in most of the state-of-the-art physics engines. As a result, a lot of data-driven approaches have been proposed based on either learning these interaction models entirely from data or augmented with prior physical knowledge~\cite{ajay2018augmenting, kloss2017combining, bauza2017probabilistic}. However, the dynamics of pushing interaction is affected by the physical attributes of the object being pushed (e.g., their shape, mass, size, etc.). As a result, the models learned for a particular object may perform poorly on novel objects~\cite{bauza2019}. Motivated by this problem and to allow study of generalization to different objects, the Omnipush dataset was released recently~\cite{bauza2019}. We also draw motivation from this problem and present a technique that can generalize to different objects during a manipulation process and thus, can be used to predict the interaction with different kind of objects. With this goal, we propose a zero-shot regression technique that can generalize using contexts available from different objects. This paper focuses on evaluations through the Omnipush dataset, but we believe that the proposed method is general and can be used to study generalization for other kinds of interactions.

Zero-shot learning algorithms in machine learning are primarily focused on classification problems where either the target classes are rare or expensive to obtain, or the number of target classes is large~\cite{Wang2019ASO}. These methods assume there is a finite number of classes and may not be easily transferable for use in regression settings. The common technique for zero-shot learning is to make use of auxiliary information or semantic representations, such as object attributes~\cite{lampert2009transfer, lampert2013attribute, welinder2010caltech} and images~\cite{zablocki2019zsl}, to assist learning a model that can generalize to unseen classes. The auxiliary information is usually embedded into a latent space, and regularization has been used to make the embedded representations for each class more separable \cite{luo2018zsl}. We apply a novel regularization where the embedding function is learned according to the distance between contexts, thus maintaining an ordering where more similar contexts are embedded closer together. This allows for a continuous spectrum of contexts instead of a finite number of class prototypes.

\section{Background}\label{sec:background}
In this section, we provide some background on the relevant learning approaches that will be referred to in the rest of the paper and allows clarity for readers not familiar with these learning approaches.

\subsection{Siamese Networks}\label{subsec:SN}

A \emph{Siamese neural network} is a network architecture used for learning a feature space describing the similarity between inputs. It consists of two copies of a network which both take a unique input and compute a distance metric between the two generated feature representations representing the similarity of the two inputs~\cite{bromley1994signature}. The parameters of the two copied networks are shared, ensuring that inputs which are similar, according to an application-specific definition, result in a lower distance.

Siamese networks have been used for object tracking, one-shot image classification and image matching \cite{bertinetto2016fully, koch2015siamese, melekhov2016siamese} and in robotics applications such as robotic surgery and indoor navigation \cite{ye2017self, yeboah2018autonomous}.

\subsection{Neural Network Masking}\label{subsec:masking}

\begin{figure}
    \centering
    \includegraphics[width=2in]{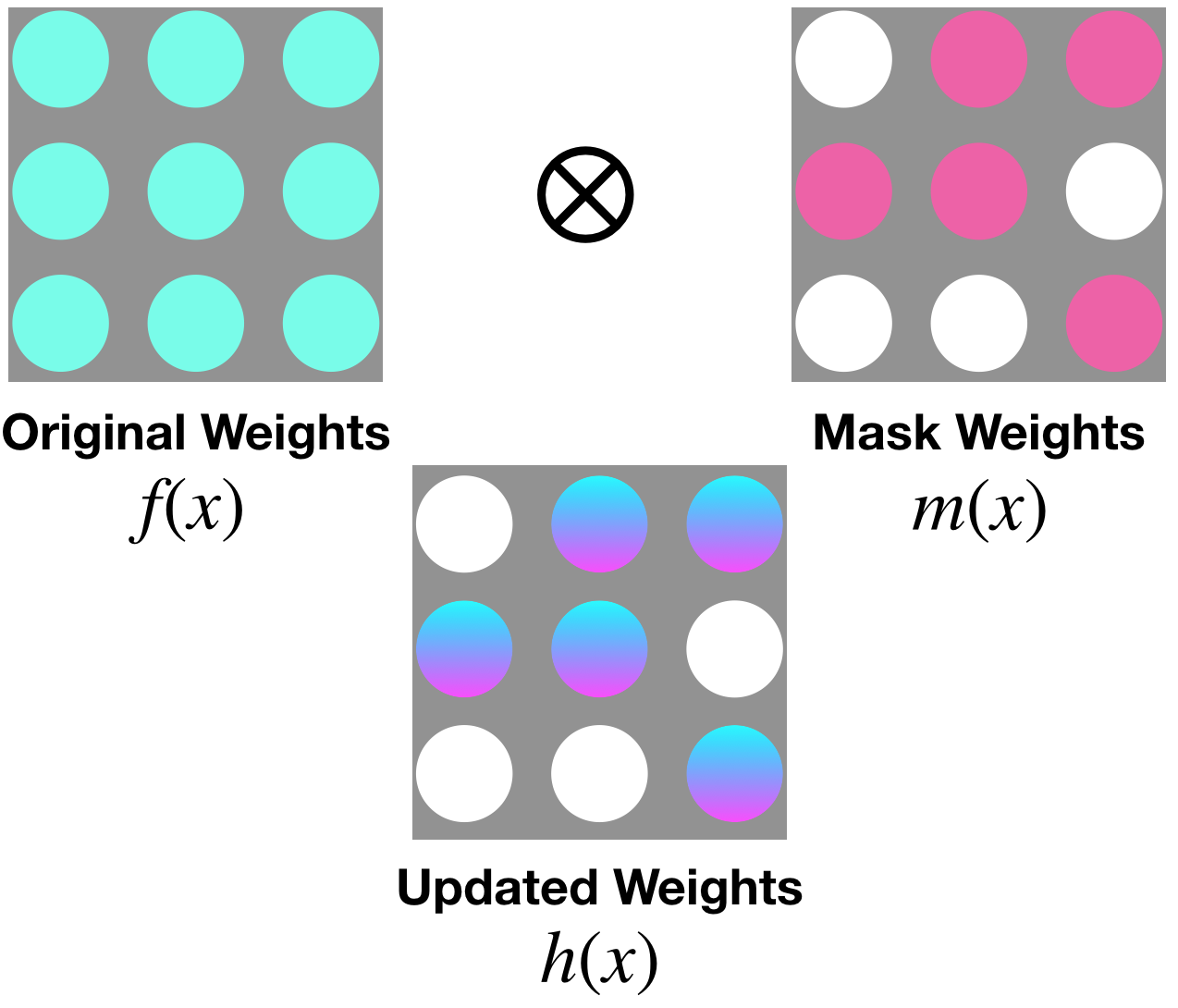}
    \caption{Overview of masking a deep neural network from \cite{mallya2018}. The original weights of the network $f(\cdot)$ are updated by a learned mask $m(\cdot)$ through elementwise product to obtain a new set of weights for the network $h(\cdot)$, allowing the network to specialize for a new set of inputs different from the original.}
    \label{fig:mask}
\end{figure}

A popular method for training deep neural networks to perform well on a new task is fine-tuning, which takes a pre-trained network and re-trains the weights so that the network performs well on the new task. A major limitation of the fine-tuning approach is that the network may no longer perform well on the original task \cite{french1999catastrophic}. Recent works demonstrate an alternative approach where a mask $m$ is learned to update the parameters of the network instead. In particular, by considering a fixed backbone $L$ layer network $$f(x) = f_L(f_{L-1}(\dots f_1(x)\dots))$$ trained on one dataset, and training an additional set of mask weights $m_1(x), \dots m_L(x)$ on a second dataset, the resulting new network 
$h(x) = h_L(h_{L-1}(\dots  h_1(x) \dots))$ where $h_i(x) = f_i\bigotimes m_i(x) $ as illustrated in Figure~\ref{fig:mask} has been found to achieve state-of-the-art performance on a second task while maintaining performance on the first~\cite{mallya2018, mancini2018adding, masana2020ternary}. We denote the elementwise product as $\bigotimes$. In general,  deep neural networks are often highly over-parametrized \cite{dauphin2013big} and a large number of weights or layers are redundant and can be pruned \cite{zhou2019deconstructing} or fine-tuned for better performance. This pruning is typically done with a binary mask \cite{zhou2019deconstructing, frankle2018lottery}, and it is shown that learning a binary mask is sufficient for state-of-the-art accuracy \cite{zhou2019deconstructing, malach2020proving,ramanujan2020s} in some cases even without training the weights of the network.  

In this work, we utilize the pairwise structure of Siamese neural networks to learn masks to better incorporate object contexts into neural network models, such that existing models can better generalize over object properties. By learning from pairwise samples of objects, we learn to incorporate the physical properties of objects such that similar objects are in turn pushed similarly because the model parameters are similar. 

\section{Proposed CAZSL}

\begin{figure}[t]
	\centering
	\includegraphics[width=\linewidth]{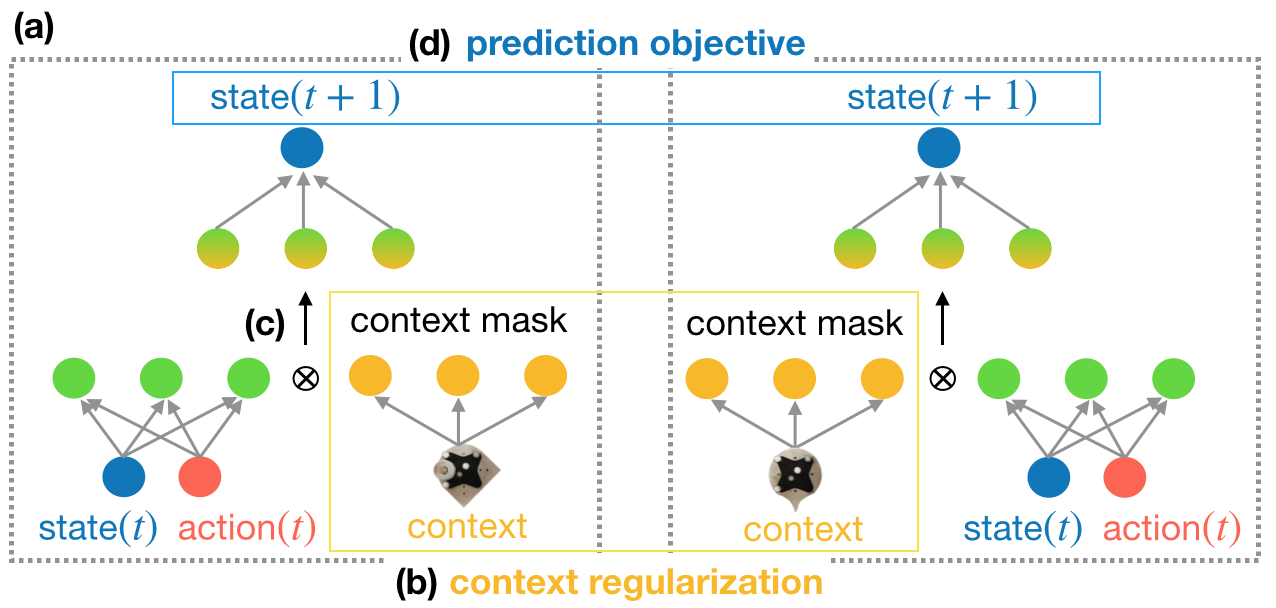}
    \vspace{-4pt}
	\caption{Proposed CAZSL model; (a) the CAZSL network is a Siamese network which ensures that the same object-push input pairs attain the same predicted state output, (b) regularization on the input context and context embedding mask enforces similar objects to have similar intermediate representations, (c) intermediate representations are altered by the context mask so that the network can guide its output based on knowledge of object properties, (d) the final predicted state is estimated from the masked intermediate representation which incorporates the context. }
	\label{fig:set-up}
\end{figure}

In this section, we describe the CAZSL method to effectively incorporate context information into the learning paradigm of neural network models. This allows the learning agents to adapt their model of a real-world physical system based on properties of the interacting objects such as mass and shape, and hence be able to generalize their predictions even towards new unseen objects. The proposed method uses a Siamese network and masking as shown pictorially in Figure~\ref{fig:set-up}. Regularization on the context inputs as well as the context mask embedding aims to enforce similar intermediate representations based on similarity in contexts. This idea is explained in more detail in the following text. 

A predictive model takes the form of $y_{t+1} = g(x_t; \Theta)$
for a deep neural network model $g$ parameterized by $\Theta$.
The inputs at time $t$ are observations $x_t$. The outputs are denoted $y_{t+1}$, which are the prediction targets at time $t+1$. 
However, the $\Theta$ learned tends to be biased towards training samples available and the resulting model does not generalize well to out-of-distribution samples. We use different symbols to denote inputs and outputs for generality, although they can contain the same entities. 

We learn the model $g$ in an end-to-end fashion to incorporate the ability to generalize to new objects through a novel combination of context mask with a regularization term.
We propose learning
\begin{equation*}
    y_{t+1} = \tilde{g}\left( x_t, c; \tilde{\Theta} \right)
\end{equation*}
where $\tilde{g}(x_t, c) = g_\ell(g_{\ell-1}(\dots g_1(x_t) \bigotimes m(c) \dots))$ is the original $\ell$-layered deep neural network with an additional non-linear context mask $m(c)$ which depends on the context $c$. The context mask is jointly learned by a neural network, and is applied as an elementwise product on the activations from the first layer. The mask augments the embedded input based on context.

Further, we encourage learning $\tilde{\Theta}$ such that if $$d(c, c') < d(c, c''),$$ then $$\|m(c) - m(c')\|_F < \|m(c) - m(c'')\|_F$$ for contexts $c$, $c'$ and $c''$ where $\|\cdot \|_F$ denotes the Frobenius norm, and $d$ is a suitably chosen distance function defining the magnitude of difference between two contexts. The key idea is that physical dynamics are more similar under more similar contexts. For instance, we would expect the objects in Figure~\ref{fig:obj0} and \ref{fig:obj1} to behave more similarly to each other when pushed than  Figure~\ref{fig:obj0} and \ref{fig:obj2}, since the first pair of objects shares three common sides while the second pair shares two.
The constraint would allow the model to generalize to new out-of-distribution contexts not in the training set, by interpolating or extrapolating based on object attributes observed in the training set. We impose this constraint through the regularization component which we refer to as \emph{context regularization}:
\begin{equation*}
    \lambda_1\left(\|m(c) - m(c')\|_F - \lambda_2 d(c, c')\right)^2
\end{equation*}
to be added to the prediction loss in the objective function. 

The twin network architecture of Siamese networks allows pairwise comparison of inputs.
The network $g$ is trained through a Siamese network structure to optimize the objective function over pairs of inputs $q^{(i)} = \paren{x^{(i)}, c^{(i)}}$ and $q^{(j)} = \paren{x^{(j)}, c^{(j)}}$, dropping time indices $t^{(i)}$ and $t^{(j)}$ in the expression for simplicity.
The complete loss function for a pair of inputs is:
\begin{align}
   & \mathcal{L}\paren{q^{(i)}, q^{(j)}; \tilde{\Theta}} =\\ 
   & \frac{1}{2}\paren{\tilde{L}\paren{q^{(i)}; \tilde{\Theta}} + \tilde{L}\paren{q^{(j)}; \tilde{\Theta}}} \nonumber\\ 
   &\hspace{1cm} + \lambda_1\left(\norm{m\paren{c^{(i)}; \tilde{\Theta}} - m\paren{c^{(j)}; \tilde{\Theta}}}_F \right. \nonumber \\ 
   &\hspace{1cm} - \left. \lambda_2 d\paren{c^{(i)}, c^{(j)}}\right)^2 \nonumber
\end{align}
where $\tilde{L}$ is the prediction loss function. 

Throughout our experiments, we model $$p\left(y_{t+1}| q_t,  \tilde{\Theta}\right) = \mathcal{N}\left( g\left( x_t, c; \tilde{\Theta} \right); \sigma^2\right)$$ and $\tilde{L}$ is the negative log likelihood of the prediction. Additionally, we consider two distance functions over the vectorized context inputs for our context regularizer:
\begin{enumerate}
    \item $L_2$ regularization: Euclidean distance function $$d\paren{c^{(i)}, c^{(j)}} = \norm{c^{(i)} - c^{(j)}}_2,$$
    \item neural regularization: kernel distance function $$d\paren{c^{(i)}, c^{(j)}} = \phi^T\left(c^{(i)}\right)\phi\left(c^{(j)}\right).$$
\end{enumerate}{}
In the kernel distance function, $$\phi(x) = W_2 \max(0, W_1x)$$ is a two layer fully-connected network, or $$\phi(x) = W \text{avg-pool} \left(\max\left(0, \text{conv}(x) \right)\right)$$ which involves learning the spatial features of $x$ through a convolutional neural network when $x$ is an image. Using $L_2$ regularization is reasonable when the context variables are continuous, and neural regularization may be more advantageous when the relationship between the context variables are highly non-linear, as in many dynamical systems. Another benefit of the neural regularization is that hyperparameter $\lambda_2$ can be absorbed and learned, and we fix  $\lambda_2 = 1$ for all experiments with neural regularization which is equivalent to not setting the second hyperparameter.

\section{Experiments and Results}
We present results to clarify, motivate and justify the use of the proposed CAZSL\footnote{\url{https://www.merl.com/research/license/CAZSL}} method for zero-shot learning. To do so, we perform a series of numerical experiments to answer the following questions.
\begin{enumerate}
    \item Is the inclusion of context helpful towards learning?
    \item Does CAZSL improve regression performance on out-of-distribution samples?
    \item How should the distance function in CAZSL be selected?
\end{enumerate}

We evaluate our method on a simple regression task as well as six experiments using two contexts from the Omnipush dataset~\cite{bauza2019}. 
In the following subsections, we abbreviate competing methods evaluated as ANP (attentive neural process), FCN (fully-connected network), and FCN + CC (FCN with context concatenated to input). ANP is a meta-learning method that uses an attention mechanism on relevant context points for regression~\cite{kim2019attentive}, and FCN is a 4-layer fully-connected neural network. These two methods are used in the Omnipush data-release paper~\cite{bauza2019} and they do not make use of context information. 
We apply our proposed context mask and regularization directly on FCN for easy comparison of their effects.
We abbreviate variations of our proposed CAZSL method for ablation studies as FCN + CM (FCN with context mask), FCN + CM + L2Reg (FCN + CM with $L_2$ context regularization), and FCN + CM + NeuralReg (FCN + CM with neural context regularization).

We point out that the FCN predicts a Gaussian density for each sample as defined by a mean $\hat{y}_{t+1}$ and standard deviation $\hat{\sigma}$. The mean parameter is evaluated by root-mean-square error (RMSE). We also report the standard deviation (STD) to give a sense of the prediction uncertainty. All values reported correspond to test performance with parameters from the last training epoch. 

\textbf{Hyperparameters}: 
For the simple regression task in Section \ref{subsec:regression}, we train all models for $500$ epochs with the Adam optimizer using a learning rate of $\eta = 0.002$, and a batch size of $32$. This configuration is sufficient for convergence due to the small size of the simulation dataset.

For experiments on the Omnipush dataset in Section \ref{subsec:omnipush}, we use the same configurations as in \cite{bauza2019}, that is,
we train all models for a maximum of $3000$ epochs with the Adam optimizer using a learning rate $\eta = 0.002$, and a batch size of $64$. The ANP model in \cite{bauza2019} is trained for $5000$ epochs with warm-up step of $4000$ whereas we train for $3000$ epochs with warm-up step of $2000$ to match the number of epochs of all other methods. Our replicated results of methods in \cite{bauza2019} are comparable with the original results reported.

\subsection{Regression}
\label{subsec:regression}

We use a simple regression task to illustrate the effects of the proposed context mask and regularization on FCN.
We simulate univariate Gaussian processes with the RBF kernel:
\begin{equation*}
    K(a,b) = \xi^2 \exp\paren{-\frac{\|a-b\|^2}{2\ell}}
\end{equation*}
where $\xi$ controls the scale and $\ell$ is the bandwidth controlling how far the data can be extrapolated. The parameters are drawn uniformly at random as $\xi \sim \texttt {Unif}(0.1, 10)$ and $\ell \sim \texttt {Unif}(0.1, 10)$. The training set consists a total of 4000 samples extracted from Gaussian processes generated with 200 parameter sets. 20 samples are extracted per parameter set, and denoting $z_t$ as the observation at time $t$, each sample has predictor $x_t = \{z_{t-2}, z_{t-1}, z_t\}$ which is a subsequence of 3 historical observations and response $y_{t} = \{z_{t+1}\}$ as the predictive target. The context variable is the kernel parameters $c = \{\xi, \ell\}$. The test set consists 400 out-of-distribution samples corresponding to 20 new parameter sets.

The simulation is repeated 10 times. We set hyperparameters $\lambda_1 = 0.0001$ and $\lambda_2 = 10$ for the applicable models. We use a small degree of regularization since this regression task is relatively simple and this set of hyperparameters gives good training performance. 
From Table~\ref{tab: regression}, all variations of our proposed method outperform the baselines FCN and FCN + CC on the test set. 
We note that context concatenation has decreased accuracy while context masking has improved accuracy, which reflects the effectiveness of masking in the embedding space. 
The use of regularization further improves performance, and FCN + CM + L2Reg has the largest $9.26\%$ reduction of RMSE over FCN.

\begin{table}[]
\centering
\begin{tabular}{|l|l|l|}
\hline
                                                                & RMSE           & STD            \\ \hline
FCN                                                             & 0.108          & 0.131          \\ \hline
\begin{tabular}[c]{@{}l@{}}FCN + CC\end{tabular}                & 0.146          & 0.133          \\ \hhline{|=|=|=|}
\begin{tabular}[c]{@{}l@{}}FCN + CM\end{tabular}                & 0.105          & 0.121          \\ \hline
\begin{tabular}[c]{@{}l@{}}FCN + CM + L2Reg\end{tabular}        & \textbf{0.098} & 0.100          \\ \hline
\begin{tabular}[c]{@{}l@{}}FCN + CM + NeuralReg\end{tabular}    & \underline{0.103}          &  0.113         \\ \hline
\end{tabular}
\caption{Average one-step prediction performance on out-of-distribution Gaussian processes samples across 10 simulations. RMSE and STD are based on mean and standard deviation estimates in the Gaussian log-likelihood objective function respectively. \label{tab: regression}}
\end{table}

\subsection{Omnipush Dataset}
\label{subsec:omnipush}

\begin{figure}[t]
	\centering
	\includegraphics[width=0.90\linewidth]{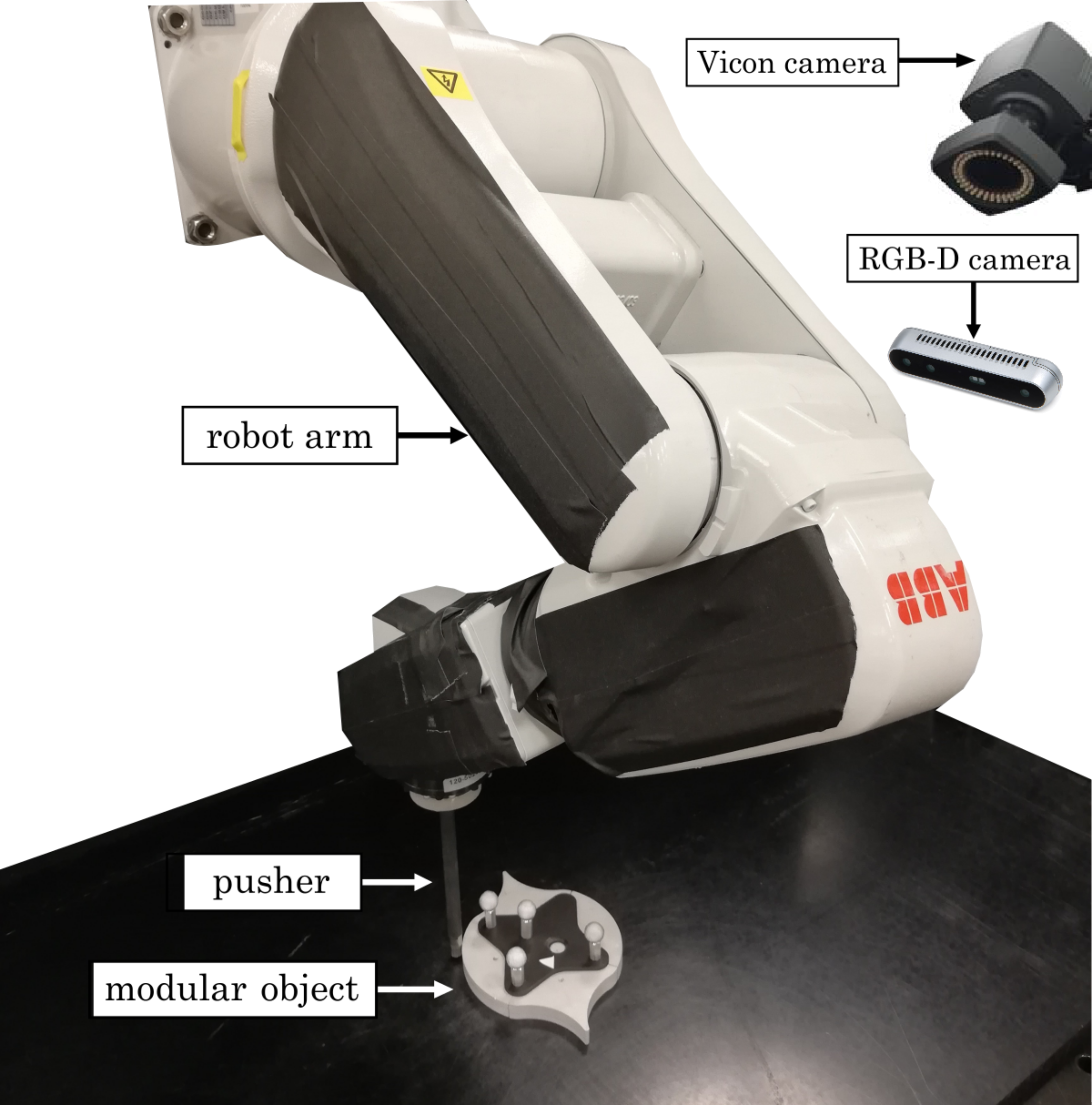}
	\caption{Omnipush dataset collection setup. The data is collected using an ABB industrial arm. The pusher is a steel rod attached to the end-effector of the arm. This steel arm interacts with the objects which are pushed during the experiments. The picture is reproduced with permission from~\cite{bauza2019}.}
	\label{fig:exp_setup}
\end{figure}

\textbf{DataSet Description:}
The Omnipush dataset~\cite{bauza2019} collected 250 pushes per object for 250 objects on ABS surface (hard plastic). 
The data collection setup for pushing is shown in Figure~\ref{fig:exp_setup}. 
The objects are constructed to explore key factors that affect pushing -- the shape of the object and its mass distribution -- which have not been broadly explored in previous datasets and allow for study of generalization in model learning. 
Each side of the object has four possible shapes (concave, triangular, circular, rectangular) with three types of extra weights (0g, 60g, 150g). The triangular shape allows two positions (interior, exterior) for extra weights to be attached. A maximum of two weights are attached per object. We denote the shape and mass distribution of the objects as context, and experiment with two types of context variables:
\begin{enumerate}
    \item Indicator context: length--36 binary vector indicating the shape, extra weight and its position for each side
    \item Visual context: numerical array representing top-down view of object displayed in Figure~\ref{fig:data_sample}.
\end{enumerate}

This allows us to test the generalization capability of our proposed CAZSL technique. The visual context is $32\times 32$, resized from an original $481\times 481$ image\footnote{We resize original images in~\cite{bauza2019} using the resize function with default parameters in scikit-image.}. The dataset further has 250 pushes per object for 10 objects on plywood surface. 
More details of the dataset can be found on the website \url{http://web.mit.edu/mcube/omnipush-dataset/} and the corresponding paper~\cite{bauza2019}. 

The prediction task is to estimate the ending location and orientation of the object after being pushed.
In data collection, the pusher is set to move at constant speed. Treating the location and angle of the object as the origin $\paren{x^{(o)}_t, y^{(o)}_t, a^{(o)}_t} = (0,0,0)$, the model input is $\paren{x^{(p)}_t, y^{(p)}_t, a^{(p)}_t}$ containing the location and angle of the pusher with respect to the object. As in \cite{bauza2019}, the location and angle of the object are not needed as input since they are constant at zero with this perspective.
The model output is the 3-dimensional vector $\paren{x^{(o)}_{t+1}, y^{(o)}_{t+1}, a^{(o)}_{t+1}}$.
To give a more intuitive representation of model accuracy, we convert RMSE to millimeters by multiplying it by 21.92mm, as done by the authors in \cite{bauza2019}.

\textbf{Experiment Setups:}
We use three setups to evaluate generalization performance of models across objects,
\begin{enumerate}
    \item \emph{Different objects}: training and test objects have different characteristics, that is, the combination of shapes, weights and weight positions for four sides;
    \item \emph{Different surfaces}: training objects are pushed on ABS surface and test objects are pushed on plywood;
    \item \emph{Different weights}: training and test objects have a different number of extra weights attached.
\end{enumerate}
The \emph{Different surfaces} setup allows evaluation for generalization performance beyond the provided context since surface information is not provided during training. We note that some objects pushed on plywood do not have images provided, and hence we only use indicator context for this setup. 

The \emph{Different weights} setup is further split into three sub-setups. There are three possible number of weights $\{0,1,2\}$ per object, and we use objects in each of the three options in turn as test objects, and the remaining objects as training objects.

For all experiments with indicator context, we set CAZSL hyperparameters $\lambda_1 = 0.01$ and $\lambda_2 = 10$ where applicable. For visual context, we set $\lambda_1 = 0.01$ and $\lambda_2 = 0.01$ where applicable. The smaller $\lambda_2$ is to balance the higher dimensions of the visual context variables. When neural regularization is used, we treat $\lambda_2 = 1$ which is equivalent to not having the second hyperparameter. For context concatenation and context masking, visual contexts are first embedded by a two-layer convolutional neural network learned jointly with the rest of the network.

\textbf{Results:}
From Table~\ref{tab:diff_objnsurface} and \ref{tab:diff_wt}, CAZSL models consistently have improved performance over the baseline FCN and FCN + CC, reflecting that the inclusion of contextual information helps learning but the context should be applied to the embedding space instead of directly concatenated in the observation space. The RMSE of ANP is consistently between 0.22 and 0.28. Since ANP is a meta-learning method which is aimed at object-generalization, we would expect its performance to be fairly consistent across setups. In comparison with ANP, our CAZSL models achieved better performance except in two sub-cases in Table~\ref{tab:diff_wt_i} for learning \emph{Different weights} with indicator context. However, we note that our $L_2$ regularized approach outperforms the ANP in the $0$ weight test set of the \emph{Different weights} experiment, indicating the ability to better generalize to unknown mass distributions. Moreover, with the use of visual contexts which contain more detailed contextual information, CAZSL models consistently outperform ANP. As expected, visual contexts result in better performance than indicator contexts since the former provides more fine-grained comparison between object shapes.

Comparing between the $L_2$ and neural regularizations in CAZSL models, we see that the former has lower RMSE when indicator context is used. Since the indicator context is a sparse binary vector, the neural network used for its embedding in neural regularization is possibly over-parameterized, hence resulting in overfitting. When visual context is used, the performance difference between the two choices of regularization is marginal. The convolutional neural network used for context embedding is able to extract spatial features possibly relating to the object geometry and mass distribution, and hence the kernel distance function learned is able to better discern differences between visual contexts. These results suggest that neural regularization might become more suitable with more complex or higher-dimensional context variables. 

In summary, in all experiments, we find that our CAZSL models outperform baseline counterparts which do not implement context masking and regularization, and perform comparably or better than the ANP meta-learning baseline. We also observe that using indicator contexts improves performance over using no context most of the time, and using visual contexts improves performance over using indicator contexts or no context in all experiments. This suggests that increasing details in contextual information can be utilized to help learning.

\begin{table*}[h]
    \centering
    
    \begin{subtable}{\linewidth}
      \centering
      
        \begin{tabular}{| c | c | c | c | c | c | c |}
        \hline
        & \multicolumn{6}{c|}{Different objects} \\
        & \multicolumn{3}{c|}{Indicator context} & \multicolumn{3}{c|}{Visual context} \\ 
                            & RMSE          & STD       & Dist. (mm)        & RMSE      & STD       & Dist. (mm)       \\ \hline
        ANP                 & 0.222         & 0.079     & 4.87              & 0.222     & 0.079     & 4.87 \\ \hline
        FCN                 & 0.330         & 0.149     & 7.23              & 0.330     & 0.149     & 7.23\\ \hline
        FCN + CC            & 0.224         & 0.040     & 4.91              & \underline{0.205}     & 0.063     & 4.49 \\ \hhline{|=|=|=|=|=|=|=|}
        FCN + CM            & \underline{0.210}         & 0.043     & 4.60              & \textbf{0.193}     & 0.039     & 4.23 \\ \hline
        FCN + CM + L2Reg    & \textbf{0.205}         & 0.029     & 4.49              & \textbf{0.193}     & 0.060     & 4.23 \\ \hline
        FCN + CM + NeuralReg& 0.220        &  0.037     & 4.82              & \textbf{0.193}     & 0.055     & 4.23 \\ \hline
        \end{tabular}
        \caption{Different objects: Training and test samples are from objects with different characteristics i.e. shape and mass distribution.\label{tab:diff_obj}}
        
    \end{subtable}%
    
    \begin{subtable}{\linewidth}
      \centering
      
        \begin{tabular}{| c | c | c | c |}
        \hline
        & \multicolumn{3}{c|}{Different surfaces: Indicator context} \\
                            &  RMSE     & STD       & Dist. (mm) \\ \hline
        ANP                 & 0.271     & 0.064     & 5.94          \\ \hline
        FCN                 & 0.328     & 0.154     & 7.19          \\ \hline
        FCN + CC            & 0.264     & 0.046     & 5.79          \\ \hhline{|=|=|=|=|}
        FCN + CM            & \textbf{0.257}     & 0.036     & 5.63          \\ \hline
        FCN + CM + L2Reg    & \underline{0.260}     & 0.035     & 5.70          \\ \hline
        FCN + CM + NeuralReg& 0.263     & 0.045     & 5.76          \\ \hline
        \end{tabular}
        \caption{Different surfaces: Training samples are from objects pushed on ABS surface (hard plastic), and test samples are from objects pushed on plywood surface. Some test objects have different characteristics from training objects.\label{tab:diff_surface}}
        
    \end{subtable}%

    \caption{Average method performance under multiple setups where test samples have out-of-distribution properties. RMSE and STD are based on mean and standard deviation estimates in the Gaussian log-likelihood objective function respectively. Setups same as in \cite{bauza2019}. \label{tab:diff_objnsurface}}

\end{table*}
    
\begin{table*}[h]
    \centering
    \begin{subtable}{\linewidth}
      \centering
      
        \begin{tabular}{| c | c | c | c | c | c | c | c | c | c |}
        \hline
        & \multicolumn{9}{c|}{Different weights: Indicator context} \\
        & \multicolumn{3}{c|}{0 Weight} & \multicolumn{3}{c|}{1 Weight}  & \multicolumn{3}{c|}{2 Weights}\\
                            &  RMSE     & STD   & Dist. (mm)    &  RMSE     & STD   & Dist. (mm)    &  RMSE     & STD   & Dist. (mm) \\ \hline
        ANP                 &  0.252    & 0.079 & 5.52          &  \textbf{0.250}   & 0.070  & 5.48          &  \textbf{0.242}    & 0.079    & 5.30\\ \hline
        FCN                 &  0.327    & 0.144 & 7.17          &  0.356    & 0.127 & 7.80          &  0.329    & 0.163 & 7.21\\ \hline
        FCN + CC            &  0.258    & 0.073 & 5.66          &  0.359    & 0.049 & 7.87          &  0.331    & 0.043 & 7.26\\ \hhline{|=|=|=|=|=|=|=|=|=|=|}
        FCN + CM            &  \underline{0.235}    & 0.038 & 5.15          &  0.267    & 0.033 & 5.85          &  \underline{0.266}    & 0.030 & 5.83\\ \hline
        FCN + CM + L2Reg    &  \textbf{0.227}    & 0.040 & 4.98          &  \underline{0.257}    & 0.034 & 5.63          &  0.272    & 0.033 & 5.96\\ \hline
        FCN + CM + NeuralReg&  0.254    & 0.039 & 5.57          &  0.294    & 0.034 & 6.44          &  0.294    & 0.036 & 6.44\\ \hline
        \end{tabular}
        \caption{Different weights: Indicator context. \label{tab:diff_wt_i}}
        
    \end{subtable}%

    \begin{subtable}{\linewidth}
      \centering
      
        \begin{tabular}{| c | c | c | c | c | c | c | c | c | c |}
        \hline
        & \multicolumn{9}{c|}{Different weights: Visual context} \\
        & \multicolumn{3}{c|}{0 Weight} & \multicolumn{3}{c|}{1 Weight}  & \multicolumn{3}{c|}{2 Weights}\\
                            &  RMSE     & STD   & Dist. (mm)    &  RMSE     & STD   & Dist. (mm)    &  RMSE     & STD   & Dist. (mm) \\ \hline
        ANP                 &  0.252    & 0.079 & 5.52          &  0.250    & 0.070    & 5.48          &  0.242    & 0.079    & 5.30\\ \hline
        FCN                 &  0.327    & 0.144 & 7.17          &  0.356    & 0.127 & 7.80          &  0.329    & 0.163 & 7.21\\ \hline
        FCN + CC            &  0.239    & 0.044 & 5.15          & 0.282     & 0.044 & 6.18          & 0.268     & 0.061 & 5.87\\ \hhline{|=|=|=|=|=|=|=|=|=|=|}
        FCN + CM            &  \underline{0.222}    & 0.034 & 4.89          & 0.230     & 0.032 & 5.04          & \underline{0.219}     & 0.039 & 4.80\\ \hline
        FCN + CM + L2Reg    &  \textbf{0.209}    & 0.079 & 4.60          & \textbf{0.220}     & 0.051 & 4.82          & \textbf{0.218}     & 0.079 & 4.78\\ \hline
        FCN + CM + NeuralReg&  \textbf{0.209}    & 0.064 & 4.56          & \underline{0.222}     & 0.050 & 4.87          & \textbf{0.218}     & 0.054 & 4.78\\ \hline
        \end{tabular}
        \caption{Different weights: Visual context.\label{tab:diff_wt_v}}
        
    \end{subtable}%
    
    \caption{Average method performance when training and test samples are from objects with different number of weights attached, out of options from 0 to 2. For example, the heading `0 Weight' means that train samples are from objects with 1 or 2 weights attached, and test samples are from objects have no weight attached. RMSE and STD are based on mean and standard deviation estimates in the Gaussian log-likelihood objective function respectively. Setup not in \cite{bauza2019}.\label{tab:diff_wt}}    

\end{table*}

\section{Conclusion and Future Work}\label{sec:conclusion}
Robotic manipulation is hard to model as the interaction dynamics are affected by complex phenomena like dry friction, contacts, impacts, etc. which are difficult to model. Furthermore, the robots are often expected to work with unknown workpieces. As such it is challenging to create models that can predict these interactions accurately over a diverse range of objects with different physical attributes. We present a zero-shot learning method CAZSL which allows us to explicitly consider the physical attributes of different objects so that the predictive model can then be easily adapted to a novel object. We introduced a novel combination of context mask and regularization that augments model parameters based on contexts and constrains the model to predict similar behavior for objects with similar physical attributes. We tested our CAZSL models on the recently released Omnipush dataset. We demonstrate empirically that CAZSL improves performance or performs comparably to meta-learning and object-independent baselines in numerous scenarios. 

In the future, we would like to further develop the algorithm and test it on much bigger and diverse interaction datasets. We would like to further investigate the proposed method for multi-step predictive error so that it could be evaluated for control of modeled interactions. Similarly, it would be interesting to test the proposed method for prediction in other physical domains~\cite{tahersima2019deep, zhu2018bayesian}.











\bibliographystyle{IEEEtran}
\bibliography{bib}

\end{document}